\documentclass[10pt,twocolumn,letterpaper]{article}

\usepackage{iccv}
\usepackage{times}
\usepackage{epsfig}
\usepackage{graphicx}
\usepackage{amsmath}
\usepackage{amssymb,paralist}
\usepackage{booktabs}

\usepackage[accsupp]{axessibility}  % Improves PDF readability for those with disabilities.

\makeatletter
\newcommand{\printfnsymbol}[1]{%
  \textsuperscript{\@fnsymbol{#1}}%
}
\makeatother
\usepackage[breaklinks=true,bookmarks=false]{hyperref}
\iccvfinalcopy % *** Uncomment this line for the final submission

\def\iccvPaperID{23} % *** Enter the ICCV Paper ID here

\ificcvfinal\fi

\begin{document}

%%%%%%%%% TITLE
\title{BERTHop: An Effective Vision-and-Language Model

 for Chest X-ray Disease Diagnosis}
\author{Masoud Monajatipoor$^{1,3}$\thanks{equal contribution}, Mozhdeh Rouhsedaghat$^2$\printfnsymbol{1}, Liunian Harold Li$^1$,\\
Aichi Chien$^4$, C.-C. Jay Kuo$^2$, Fabien Scalzo$^1$ \& Kai-Wei Chang$^1$\\

$^1$Computer Science Department, UCLA\\
$^2$Department of Electrical Engineering, USC\\
$^3$Department of Electrical Engineering, UCLA\\
$^4$Department of Radiology, UCLA

\\\texttt{\small monajati@ucla.edu,rouhseda@usc.edu,aichi@ucla.edu,jckou@usc.edu}\\ \texttt{\{\small liunian.harold.li,fab,kwchang\}@cs.ucla.edu}}
%\author{Masoud Monajatipoor\\
%Institution1\\
%Institution1 address\\
%{\tt\small firstauthor@i1.org}
% For a paper whose authors are all at the same institution,
% omit the following lines up until the closing ``}''.
% Additional authors and addresses can be added with ``\and'',
% just like the second author.
% To save space, use either the email address or home page, not both
%\and

\maketitle
% Remove page # from the first page of camera-ready.
\ificcvfinal\thispagestyle{empty}\fi

%%%%%%%%% ABSTRACT
\begin{abstract}
   Vision-and-language (V\&L) models take image and text as input and learn to capture the associations between them. Prior studies show that pre-trained V\&L models can significantly improve the model performance for downstream tasks such as Visual Question Answering (VQA). However, V\&L models are less effective when applied in the medical domain (e.g., on X-ray images and clinical notes) due to the domain gap. In this paper, we investigate the challenges of applying pre-trained V\&L models in medical applications. In particular, we identify that the visual representation in general V\&L models is not suitable for processing medical data. To overcome this limitation, we propose BERTHop, a transformer-based model based on PixelHop++ and VisualBERT, for better capturing the associations between the two modalities. 
   %Experiments on the OpenI dataset show that BERTHop achieves an average AUC of 98.23\% on the thoracic disease diagnosis benchmark, outperforming the previous comparable method by 14.37\%. BERTHop also performs 1.73\% higher than state-of-the-art, while it is trained on a 9× smaller dataset.
   Experiments on the OpenI dataset, a commonly used thoracic disease diagnosis benchmark, show that BERTHop achieves an average Area Under the Curve (AUC) of 98.12\% which is 1.62\% higher than state-of-the-art (SOTA) while it is trained on a 9× smaller dataset.
  
   %In this paper, we discuss the challenges when applying V\&L models in medical domain applications, and propose a new approach, BERTHop, to better capture the association between two modalities. Experiments on the OpenI dataset show that our method achieves state-of-the-art performance on the thoracic disease diagnosis benchmark.
\end{abstract}

\section{Introduction}

Computer-Aided Diagnosis (CADx) \cite{giger2008computer} systems could provide valuable benefits for disease diagnosis including but not limited to improving the quality and consistency of the predictions and reducing medical mistakes as they are not subject to human error. Although most existing studies focus on diagnosis based on medical images such as chest X-ray (CXR) images \cite{ayan2019diagnosis,allaouzi2019novel,abiyev2018deep}, the radiology reports often contain substantial information (e.g. patient history and previous studies) that are difficult to be detected from the image alone. Besides, diagnosis from both image and text is more closely aligned with disease diagnosis by human experts. Therefore, V\&L models that take both images and text as input can be potentially more accurate for CADx and several attempts have been made in this direction \cite{wang2018tienet,zhang2019text,li2020comparison}.
%Therefore, V\&L models that take both images and text as input can be potentially more accurate for CADx. 
 %this research area is not explored enough.
% who use both image and radiology report. 
%In particular, the radiology reports often contain substantial information that are hard to be detected from the image alone. 
%Recently, a few attempts have been made to extract information from both medical images and text reports using V\&L models \cite{wang2018tienet,zhang2019text,li2020comparison}
%In this paper, we study how to leverage V\&L models for helping CADx.

%Recently, a few attempts have been made to extract information from both medical images and text reports \cite{wang2018tienet,zhang2019text,li2020comparison}. To understand their limitations, We first discuss their limitations in Section \ref{related}.
%Although some recent works proposed models for disease diagnosis based on only medical images \cite{ayan2019diagnosis,allaouzi2019novel,abiyev2018deep}, diagnosis from both image and text has not been explored enough while it is a closer scenario to disease diagnosis by human experts who use both image and radiology report. 
%In particular, the radiology reports often contain substantial information that are hard to be detected from the image alone. Therefore, some V\&L models have been proposed to tackle this problem \cite{wang2018tienet,zhang2019text,li2020comparison}.

\begin{figure}
  \centering
  %\fbox{\rule[-.5cm]{0cm}{4cm} \rule[-.5cm]{4cm}{0cm}}
  %\includegraphics[width=10cm, height=6cm]{./fig/fig2_3}scale=1.5
  \includegraphics[width=\columnwidth]{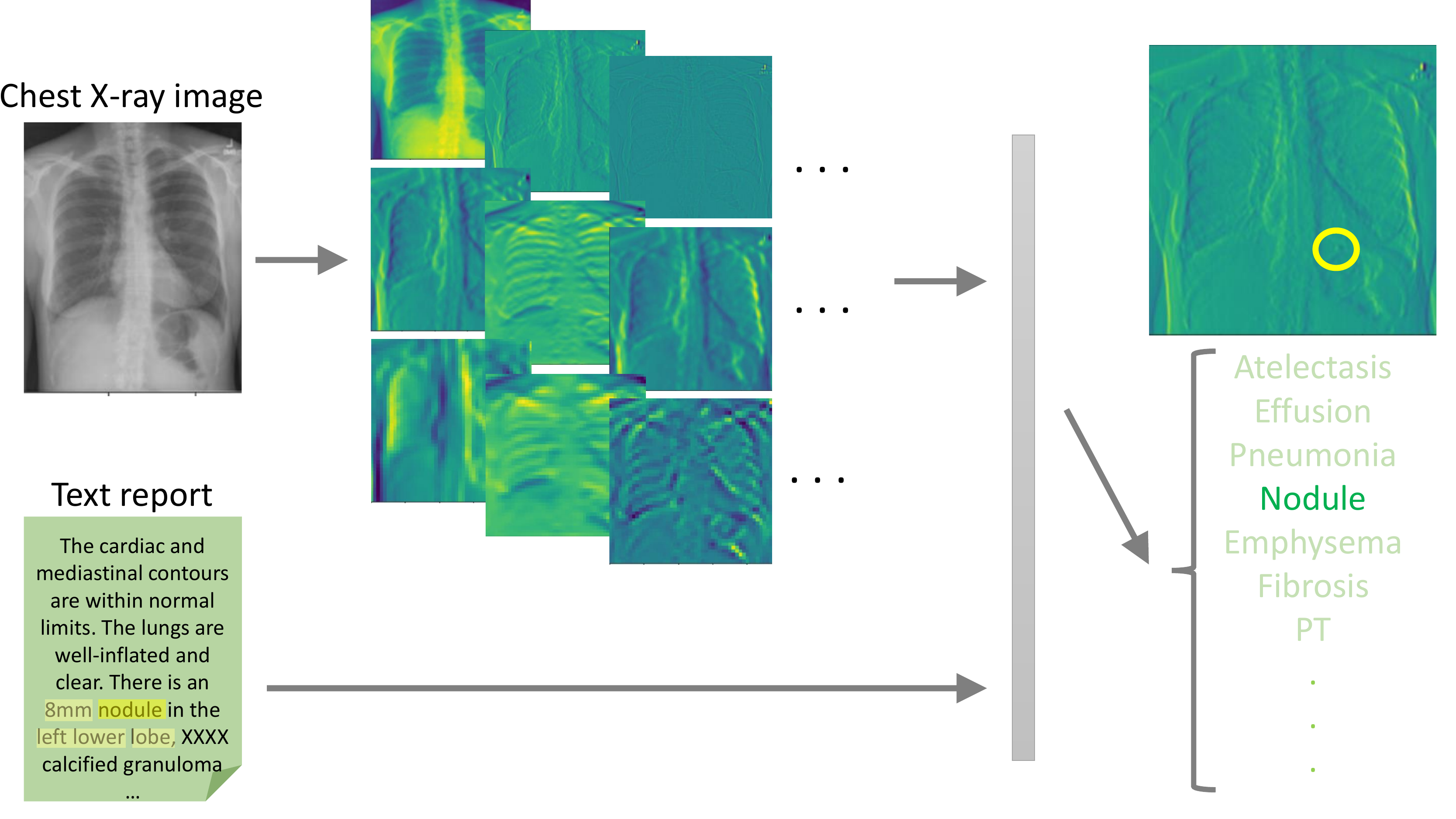}
  \caption{An overview of BERTHop. BERTHop takes X-ray image and clinical report as input. It first encodes the image and text and extracts potential features from both modalities. Then a transformer-based model learns the associations between these two modalities. By applying appropriate vision and text extractor, the model is capable to identify the abnormality and associate it with the text labels.}
\label{fig:stat1}
\end{figure}

%V\&L models integrate these two modalities for a certain task. 
However, the shortage of annotated data in the medical domain makes utilizing V\&L models challenging. 
%Generally, there is the problem of lack of annotated data in the medical domain. 
Annotating medical data is an expensive process as it requires human experts. 
%Besides, some challenging diagnosis may raise disagreements among experts while such pitfalls rarely exist in the general domain annotation. In these cases, we need several experts' opinions to achieve a reliable annotation. 
Although a couple of recent large-scale auto-labeled datasets have been provided for some medical tasks, e.g., chest X-ray \cite{wang2017chestx,bustos2020padchest,johnson2019mimic}, they are often noisy (low-quality) and degrade the performance of models. Besides, such datasets are not available for most medical tasks. %Therefore, the problem still remains.
%their labels are generated  these datasets are not annotated by experts. Besides, such datasets are not available for most medical tasks. 
%these datasets are often noisy. 
Therefore, training V\&L models with limited annotated data remains a key challenge. 

%and 
%not annotated by experts and they are not available for most medical tasks. The shortage of annotated data is a motivation to take advantage of pre-trained V\&L models for knowledge transfer from large-scale datasets. 

Recently, pre-trained V\&L models have been proposed for reducing the amount of labeled data required to train an accurate downstream model~\cite{li2019visualbert,tan2019lxmert,su2019vl,lu2019vilbert} in the general domain. These models are first trained on large-scale image caption data with self-supervision signals (e.g., using masked language model loss) to learn the association between objects and text tokens. Then, the pre-trained V\&L models are used to initialize the downstream models and fine-tuned on the target tasks. 
%Mozhdeh
In most V\&L tasks, it has been reported that V\&L pre-training is a major source of performance improvement. However, we identify a key problem in applying common pre-trained V\&L models for the medical domain: the large domain gap between the medical (target) and the general domain (source) makes such pre-train and fine-tune paradigm considerably less effective in the medical domain. Therefore, domain-specific designs have to be applied.

Notably, V\&L models mainly leverage object-centric feature extraction methods such as Faster R-CNN \cite{ren2016faster} which is pre-trained on general domain to detect everyday objects, e.g., cats, and dogs. However, the abnormalities in the X-ray images do not resemble everyday objects and will likely be ignored by a general-domain object detector.

To overcome this challenge, we propose BERTHop, a transformer-based V\&L model designed for medical applications.
BERTHop resolves the domain gap issue by leveraging pre-training language encoder, BlueBERT \cite{peng2019transfer}, a BERT \cite{devlin2018bert} variant that has been trained on biomedical and clinical datasets. Furthermore, in BERTHop, the visual encoder of the V\&L architecture is redesigned leveraging PixelHop++ \cite{chen2020pixelhop++} and is fully unsupervised which significantly reduces the need for labeled data \cite{ rouhsedaghat2021successive}. PixelHop++ can extract image representations at different frequency levels that is beneficial for abnormality detection.

We evaluate BERTHop by conducting extensive experiments and analysis for CADx in chest disease diagnosis on the OpenI dataset \cite{demner2016preparing}. 
The OpenI dataset contains thoracic diseases, including 14 common chest diseases. Compared to SOTA (TieNET \cite{wang2018tienet}), BERTHop outperforms in 11 out of 14 thoracic diseases diagnoses and achieves an average AUC of 98.23\% that is 1.73\% higher, using significantly less training data (TieNet is trained on the ChestX-ray14 \cite{wang2017chestx} dataset that is 9 times larger than OpenI). Compared to the similar transformer-based V\&L model pre-trained on general domain and fine-tuned on OpenI \cite{li2019visualbert,li2020comparison}, BERTHop requires no expensive V\&L pre-training yet outperforms it by 14.37\%.

We summarize our contributions as follows: (1) We propose BERTHop, a novel data-efficient V\&L model for CXR disease diagnosis surpassing existing approaches. (2) Our proposed model incorporates PixelHop++ into a transformer-based model. To the best of our knowledge, this is the first study which integrates PixelHop++ and Deep Neural Network (DNN) models. (3) We conduct extensive experiments to demonstrate the effectiveness of each submodel we used in BERTHop. (4) We study how transformer initialization with a model, pre-trained on in-domain data (even on a single modality) is highly beneficial in the medical domain. 

%The summery of our paper is as the following. In Section \ref{related}, we review the related work, and discuss proposed models and their challenges that is the motivation of our work. In Section \ref{app}, we propose our method and %??
%explain the architecture of its submodels. %present an overview of the available chest X-ray vision-language datasets and their limitations in \ref{dataset}. 
%In Section \ref{exp}, we present the experiment setup and results. Then, the ablation studies are provided in section \ref{abl}. Finally, in Section \ref{fut} and \ref{con}, we offer the future work and conclusion.

\section{Related Work}
\label{related}

\paragraph{Transformer-based V\&L models} Inspired by the success of BERT for NLP tasks, various transformer-based V\&L models have been proposed \cite{li2019visualbert,chen2020uniter,tan2019lxmert}. They generally use an object detector pre-trained on Visual Genome \cite{krishna2017visual} to extract visual features from an input image and then use a transformer to model the visual features and input sentence. They are pre-trained on a massive amount of paired image-text data with a mask-and-predict objective similar to BERT. During pre-training, part of the input is masked and the objective is to predict the masked words or image regions based on the remaining contexts. Such models have been applied to many V\&L applications \cite{zhou2020unified,lu202012,chou2020visual} including the medical domain \cite{li2020comparison}. However, the performance of these models is not satisfactory due to the domain shift between the general domain and medical domain.

\paragraph{V\&L models in the medical domain}
Various CNN-RNN-based V\&L models have been proposed for disease diagnosis on CXR.
Zhang {\em et al.} \cite{zhang2019text} proposed TNNT (Text-guided Neural Network Training) which helps a CNN model get guidance from text report embedding for a more efficient training process on V\&L data and evaluated the model on four V\&L datasets including the OpenI dataset. They showed that the text report has important information that can improve the diagnosis compared with prior vision-only models, e.g., ResNet.

TieNet is a CNN-RNN-based model for V\&L embedding integrating multi-level attention layers into an end-to-end CNN-RNN framework for disease diagnosis and radiology report generation tasks. 
TieNet uses a ResNet-50 pre-trained for general-domain visual feature extraction and an RNN for V\&L fusion. As a result, it requires a large amount of in-domain training data (ChestX-ray14) for adapting to the medical domain, limiting its practical usage. In contrast, our method achieves higher performance with very limited in-domain data.

Recently, Li {\em et al.} \cite{li2020comparison} evaluated the transferability of well-known pre-trained V\&L models by fine-tuning them on MIMIC-CXR \cite{johnson2019mimic} and OpenI. However, the pre-trained models are designed and pre-trained for general-domain, and directly fine-tuning it with limited in-domain data leads to suboptimal performance. We refer to this method as VB w/ BUTD (section \ref{subsec:main result}).
%We contacted the authors but we still cannot reproduce their results reported in the paper. We report the results in Table \ref{sample-table} based on their released code\footnote{\url{https://github.com/YIKUAN8/Transformers-VQA/blob/master/openI_VQA.ipynb}} as VB w/ BUTD (fine-tuned VisualBERT).
%It claims to report AUC scores of the model while accuracy scores have been reported based on the code\footnote{\url{https://github.com/YIKUAN8/Transformers-VQA/blob/master/openI_VQA.ipynb}}. We reimplemented the model and got the same class accuracies as reported in the paper and computed the actual AUCs that are shown in Table \ref{sample-table} as VB w/ BUTD which refers to fine-tuned VisualBERT.
%We contacted the authors but we still cannot reproduce their results reported in the paper. We report the results in Table \ref{sample-table} based on their released code \footnote{\url{https://github.com/YIKUAN8/Transformers-VQA/blob/master/openI_VQA.ipynb}} as VB w/ BUTD (fine-tuned VisualBERT). \textcolor{red}{Harold: I might rewrite as: Li {\em et al.} \cite{li2020comparison} evaluated the transferability of pre-trained models \cite{li2019visualbert,tan2019lxmert,chen2020uniter} by fine-tuning them on Mimic-CXR and OpenI. However, the pre-trained models are designed and pre-trained for general-domain V\&L modeling, direct fine-tuning with limited in-domain data leads to suboptimal performance.}

\begin{figure*}[ht]
  \centering
  %\fbox{\rule[-.5cm]{0cm}{4cm} \rule[-.5cm]{4cm}{0cm}}
  %\includegraphics[width=10cm, height=6cm]{./fig/fig2_3}scale=1.5
  \includegraphics[scale=0.4]{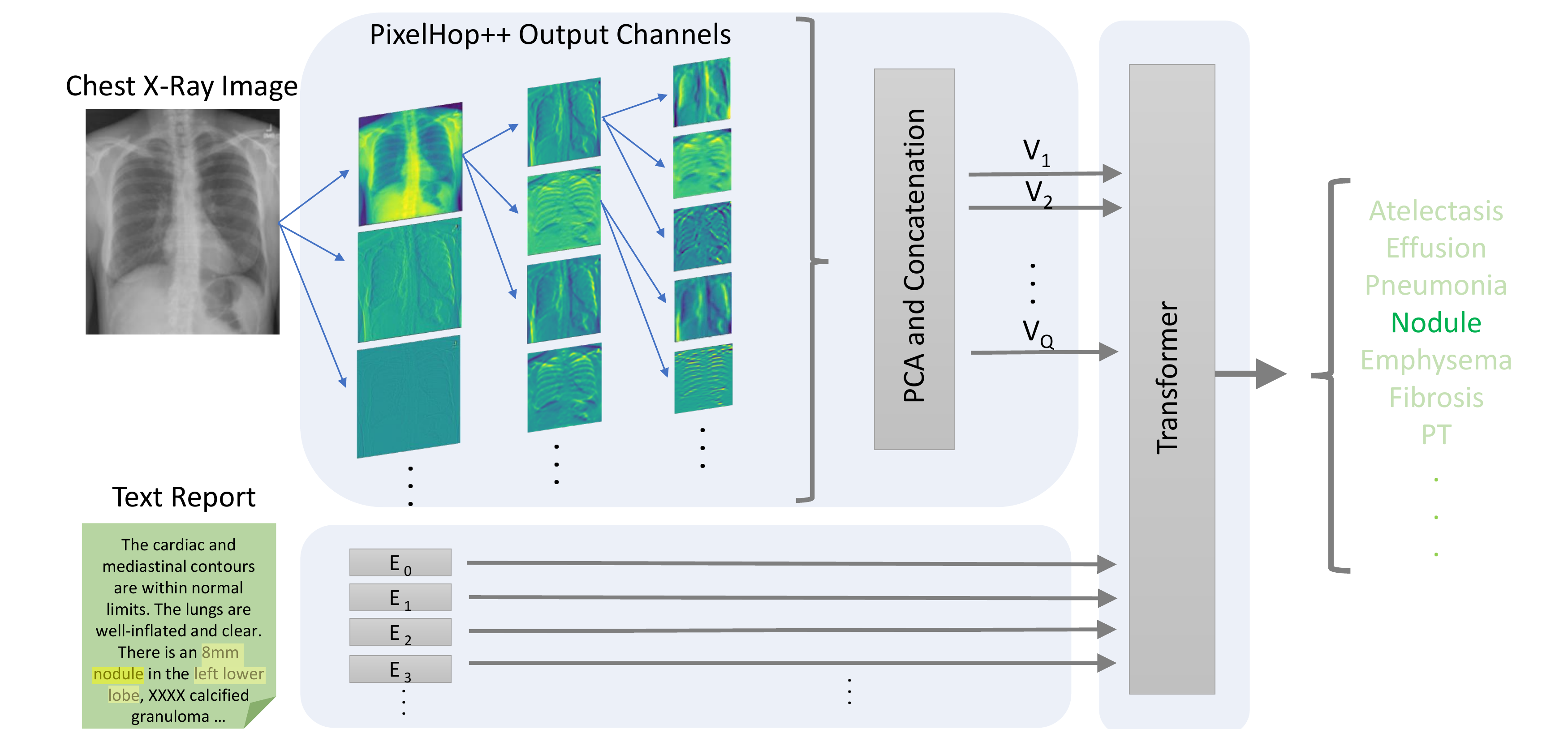}
  \caption{The proposed BERTHop framework for CXR disease diagnosis. A PixelHop++ model followed by a ``PCA and concatenation'' block is used to generate Q feature vectors. These features along with language embedding  are fed to the transformer that is initialized with BlueBERT.}
\label{fig:arch}
\end{figure*}

\paragraph{PixelHop++ for visual feature learning}
PixelHop++ is originally proposed as an alternative to deep convolutional neural networks for feature extraction from images and video frames in resource-constrained environments. It is a multi-level model which generates output channels representing an image at different frequencies.
% It usually requires an extra step of feature extraction from its output channels to generate final feature vectors. 

PixelHop++ is used in various applications and shown to be highly effective on small datasets. These applications include face gender classification \cite{rouhsedaghat2020facehop}, face recognition \cite{rouhsedaghat2020low}, and deep fake detection  \cite{chen2021defakehop}.
%FaceHop \cite{rouhsedaghat2020facehop} uses PixelHop++ for face gender classification, %It  applies Principal Component Analysis (PCA) to selected regions from PixelHop++ output channels to form final feature vectors. 
%and DefakeHop \cite{chen2021defakehop} applies PixelHop++ to face salient regions for deep fake detection from videos.
%FaceHop \cite{rouhsedaghat2020facehop} applies Principal Component Analysis (PCA) to selected regions from PixelHop++ output channels to form final feature vectors and feeds them to a Logistic Regression (LR) classifier for face gender classification. DefakeHop \cite{chen2021defakehop} uses PixelHop++ for feature extraction from face salient regions and applies PCA to its output channels to generate feature vectors. Then, feature vectors are fed into an XGBoost classifier for deep fake detection from videos. 
It has also been recently applied to a medical task. VoxelHop \cite{liu2021voxelhop} leveraged this model on 3D Magnetic resonance imaging (MRI) imaging data and could achieve superior results for Amyotrophic Lateral Sclerosis (ALS) disease classification task.

To the best of our knowledge, this is the first study which integrates PixelHop++ and DNN models. Our proposed model takes advantage of the attention mechanism to integrate visual features extracted from PixelHop++ and the language embedding.

%------------------------------------------------------------------------
%\section{Related work}
%\bold{Pre-trained V&L models:}  
%Vision language multi-modality recently got attention in the AI field. Various vision language tasks have been defined () and several models have been proposed to solve them. More recently, V&L pre-trained models () have been proposed which achieved significant improvement especially on downstream tasks. In the medical domain though, there is a big gap that has not been explored enough. Pre-trained domain specific language models () already illustrated the potential of pre-trained on the language part. But, the pre-trained vision language has not been explored enough in the medical domain yet. Some V&L DNN models () have been proposed and achieved the state-of-the-art results for both diagnosis and report generation tasks. While DNN models are reliable for diagnosis task, they are data hungry and need lots of data for the training process. As shortage of data is a major problem in the medical domain, we proposed a model that can be trained on small data by leveraging pre-trained BlueBERT for the text encoding part and pixelHop++ for the image encoding part. Our model acheived better result by fine-tuning only on OpenI dataset with 3680 data

\section{Approach}\label{app}
Inspired by the architecture of VisualBERT, our framework uses a single transformer to integrates visual features and language embeddings. The overall framework of our proposed approach is shown in Figure \ref{fig:arch}. We first utilize PixelHop++ to extract visual features from the X-ray image; then the text (a radiology report) is encoded into subword embeddings; a joint transformer is applied on top to model the relationship between two modalities and capture implicit alignments.

There are two main differences between BERTHop and previous approaches:
%BERTHop mdiffers from previous approaches in the following two aspects:
\begin{itemize}
    \item \textbf{Visual feature encoder} Considering the lack of data in the medical domain, instead of using an object detector pre-trained on a general-domain dataset, we leverage PixelHop++, an unsupervised data-efficient method, to extract visual features. As the size of the PixelHop++ output channels is relatively large to be directly fed into the transformer, we apply Principle Component Analysis (PCA) to the output channels for dimension reduction. PCA is an orthogonal linear transformation that maps the data to a new coordinate system of lower dimension so that the variation of data is better preserved. By applying PCA to the PixelHop++ output channels, we capture the most prominent features and prevent over-fitting.
    Then, we concatenate the results to generate the final visual feature vectors. (Section \ref{Vision_encoder})
    
    \item \textbf{In-domain text pre-training} Instead of resorting to computation-extensive V\&L pre-training on a general domain image-text dataset, we find in-domain text-only pre-training considerably more beneficial in our application. Thus, we use BlueBERT as the backbone for our model, a transformer pre-trained on biomedical and clinical datasets. (Section \ref{Language_encoder})
\end{itemize}

\subsection{Visual encoder}
\label{Vision_encoder}

We argue that extracting visual features from a general-domain object detector, i.e. the BUTD \cite{anderson2018bottom} approach that is dominant in most V\&L tasks, is not suitable for the medical domain. BUTD\footnote{In the following, we use the term ``BUTD'' to refer to extracting visual features from a pre-trained object detector rather than the full model from \cite{anderson2018bottom}.} takes an image and employs a ResNet-based Faster-RCNN \cite{ren2016faster} for object detection and feature extraction from each object. The detector is pre-trained on Visual Genome \cite{krishna2017visual} to detect objects in everyday scenes. Such an approach fails to detect medical abnormalities when applied to X-ray images. The reason is that the abnormalities in the image, which are of high importance for facilitating diagnosis, usually do not resemble the normal notion of an ``object'' and will likely be ignored by a general-domain object detector. %\harold{For example, in Figure \ref{}, xxxxx}. 
Further, there exists no large-scale annotated dataset for disease abnormality detection from which to train a reliable detector \cite{shin2016deep}.

%Fundamentally, these methods assume that detected objects are the only important regions of the image and the text is describing them and their relations. % Therefore, the object detection model plays an important role in these architectures. 
%In the medical domain, the text report is talking about the abnormal regions of the chest x-ray images, while detecting these abnormalities is difficult for object detection models which are pre-trained on the general domain, notably, with the lack of annotated data in the medical domain. 

%To this end

\begin{figure}
  \centering
  %\fbox{\rule[-.5cm]{0cm}{4cm} \rule[-.5cm]{4cm}{0cm}}
  %\includegraphics[width=10cm, height=6cm]{./fig/fig2_3}scale=1.5
  \includegraphics[width=\columnwidth]{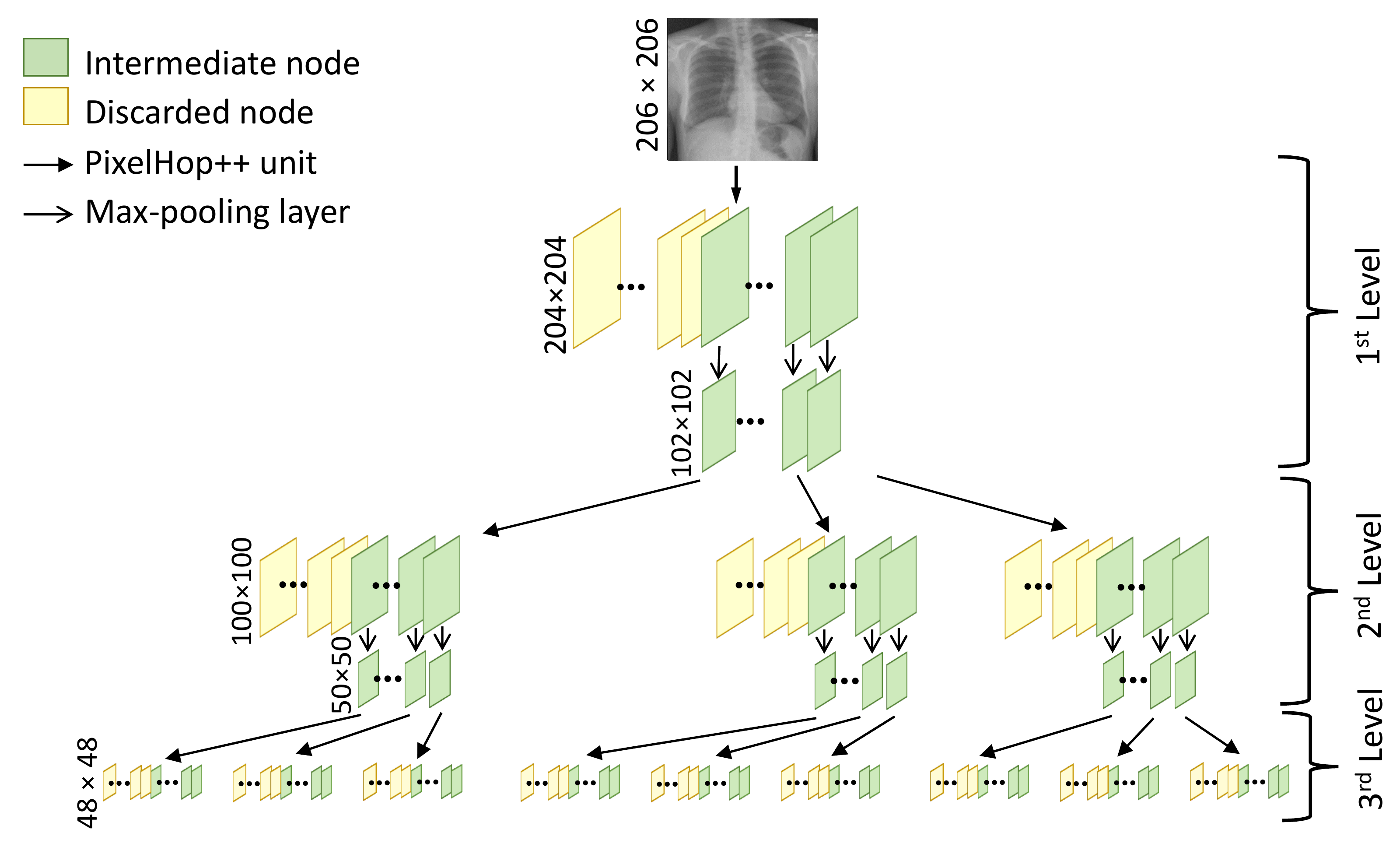}
  \caption{Data flow in a 3-level PixelHop++ model. A node represents a channel.}
\label{fig:pixel}
\end{figure}

We propose to adopt PixelHop++ \cite{chen2020pixelhop++} for unsupervised visual feature learning in the medical domain, which has been shown to be highly effective when trained on small-scale datasets. The key idea of PixelHop++ is computing the parameters of its model by a closed-form expression without using back-propagation \cite{rouhsedaghat2021successive}. As PixelHop++ leverages PCA for computing parameters, the model is able to extract image representations at various frequencies in an unsupervised manner.
Inspired by the architecture of DNN models, PixelHop++ is a multi-level model in which each level consists of one or several PixelHop++ units followed by a max-pooling layer. An illustration of data flow in a 3-level PixelHop++ model is shown in Figure \ref{fig:pixel}. When training a PixelHop++ model, parameters of PixelHop++ units (kernels and biases) are computed, and during the inference, they are used for feature extraction from pixel blocks. 

\paragraph{Training phase of PixelHop++} Suppose that we have $N$ training images of size $s_1\times s_2\times d$, where d is $1$ for gray-scale and $3$ for color images. They are all fed into a single PixelHop++ unit in the first level of the model. The goal of training a PixelHop++ unit is to compute linearly independent projection vectors (kernels) which can extract strong features from its input data. There are one or more PixelHop++ units in each level of a PixelHop++ model.

In the first step of processing data in a PixelHop++ unit, using a sliding window of size $w\times w\times d$ and a stride of $s$, patches from each training image are extracted and flattened, i.e., $x_{i1}, x_{i2}, ..., x_{iM}$ where $x_{ij}$ is the $j${th} flattened patch for image $i$, and $M$ is the number of extracted patches per image. 

In the second step, the set of all patches extracted from training images are used to compute the kernels of the PixelHop++ unit. Kernels are computed as follows:
\begin{itemize}
  \item The first kernel, called DC kernel, is the mean filter, i.e., \begin{math}\frac{1}{\sqrt{n}}\times (1, 1, ..., 1) \end{math} where n is the size of the input vector, and extracts the mean of each input vector.
  \item After computing the mean (DC component) of each vector, PCA kernels of the residuals are computed and stored as AC kernels. First, $k$ PCA kernels are the top $k$ orthogonal projection vectors which can capture the variation of residuals best.
\end{itemize}

Each image patch is projected on computed kernels and a scalar bias is added to the projection result to avoid the sign-confusion problem \cite{kuo2019interpretable}. This transformation on the input vector $(x_0, x_1, ..., x_{D-1})^T$ can be shown as follows:

\begin{equation}
\label{affine_transformation}
y_k = \sum_{d=0}^{D-1}{a_{kd}\times x_d} + {b_k}
\end{equation}
where $a_{kd}$ represents kernel parameters associated with the $k$th kernel of a PixelHop++ unit and ${b_k}$ is the kernel's corresponding bias term.

By transforming $x_{i1}, x_{i2}, ..., x_{iM}$ by a kernel in a PixelHop++ unit, one output channel is generated. For example, in the first level of the model, the PixelHop++ unit generates $1$ DC channel and $w\times w\times d - 1$ AC channels. Each channel is shown by a node in Figure \ref{fig:pixel}.

In the last step, model pruning is executed to remove the channels which include deficient data. The ratio of the variance explained by each kernel to the variance of training data is called the ``energy ratio'' of the kernel or its corresponding channel and is used as a criterion for pruning the model.
An energy ratio threshold value, $E$, is selected and model pruning is performed using the following rule:

\begin{itemize}
  \item If the energy ratio of a channel is less than $E$, it will be discarded (discarded nodes/channels in Figure \ref{fig:pixel}) as the variation of data along the corresponding kernel is very small.
  \item If the energy ratio of a channel is more than $E$, it is forwarded to the next level for further energy compaction (intermediate nodes/channels in Figure \ref{fig:pixel}).
\end{itemize}

Each output intermediate channel generated by a PixelHop++ unit will be fed into one separate PixelHop++ unit in the next level. So, except for the first level of the model, other levels contain more than one PixelHop++ unit.

\paragraph{Inference phase of PixelHop++} Data flow is similar to the training phase but all parameters including kernel weights and biases are computed during the training phase. Therefore, according to Equation \ref{affine_transformation}, feature extraction from test images is conducted in each PixelHop++ unit using the computed kernels and biases.

\subsection{In-domain text pre-training}
\label{Language_encoder}

\begin{figure}
  \centering
  %\fbox{\rule[-.5cm]{0cm}{4cm} \rule[-.5cm]{4cm}{0cm}}
  %\includegraphics[width=10cm, height=6cm]{./fig/fig2_3}scale=1.5
  \includegraphics[width=\columnwidth]{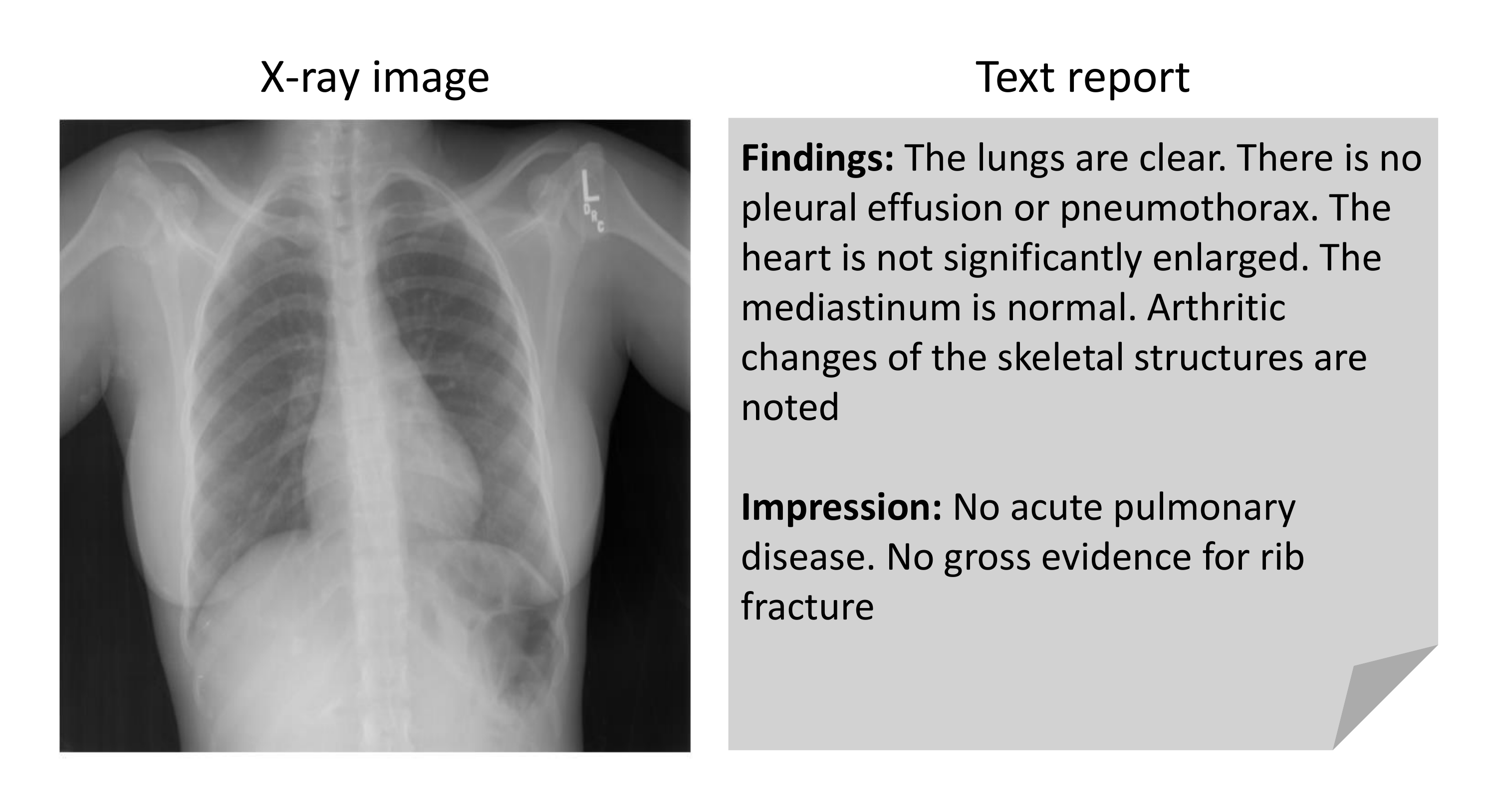}
  \caption{A sample image-text pair in the OpenI dataset. The text report from a radiologist is important for disease diagnosis but has a significantly different style compared to general-domain text.}
\label{fig:data}
\end{figure}

%In BERTHop, the text report plays an important role in guiding the transformer to pay more attention to the right visual features in the attention mechanism. %predicting the correct disease. 
As shown in an example in Figure \ref{fig:data}, the report is written by an expert radiologist, who lists the normal and abnormal observations in the ``finding'' section and other important patient information e.g. patient history %body parts, and previous studies 
in the ``impression'' section of the report. The text style of the report is drastically different from that of the pretraining corpora of BERT (Wikipedia and BookCorpus) or V\&L models (MSCOCO and Conceptual Captions).

However, previous methods \cite{li2020comparison} do not take such a significant domain gap into consideration. Rather, they initialize the transformer with a model trained on general-domain image-text corpora, as in most V\&L tasks. Meanwhile, pre-training with text-only corpora has been reported to how only marginal or no benefit \cite{tan2019lxmert}. In the medical domain, however, we find that using a transformer pre-trained on in-domain text corpora as our initialized backbone serves as a simpler yet stronger approach.

%For most V\&L tasks, it has been shown that expensive V\&L pre-training on general-domain image-caption corpora has been instrumental for achieving high-performance \cite{li2019visualbert,tan2019lxmert,chen2020uniter}. Meanwhile, pre-training with text-only corpora has been reported to how only marginal or no benefit \cite{tan2019lxmert}. In the medical domain, however, we find that in-domain text pre-training plays a more important role than pre-training on large image-caption corpora. 

%Specifically, we use BlueBERT as the backbone in our model. 
Peng {\em et al.} \cite{peng2019transfer} proposed a Biomedical Language Understanding Evaluation (BLUE) benchmark which evaluated the performance of BERT and Elmo \cite{peters2018deep} on 5 common biomedical text-mining tasks with ten corpora and showed the superiority of BERT when is pre-trained on biomedical and clinical datasets (BlueBERT).\footnote{\url{https://github.com/ncbi-nlp/bluebert}}. %They made the models and datasets with various versions publicly available.
 %?
Recently, BlueBERT has been widely used in the bioNLP community for various NLP tasks \cite{gu2020domain,fraser2019extracting,wada2020pre} and  a few V\&L tasks, e.g, data labeling \cite{jain2021visualchexbert}. Thus, we leverage this pre-trained version of BERT as the backbone in BERTHop and initialized its single-stream transformer \cite{li2019visualbert} with BlueBERT to better capture the clinical report information.

%Radiology reports usually contain richer information than its corresponding image and can boost the detection performance. It generally includes a ``Finding'' and an ``impression'' section. Radiologists list the normal and abnormal observations in the ``finding'' section and other important patient information including patient history and previous studies in the ``impression'' section of the report. A sample report is shown in Fig \ref{fig:data}.
%Therefore, we propose to initialize the transformer with BlueBERT. BlueBERT is an extension to BERT which has been trained on biomedical and clinical datasets.

%\begin{redcolor}
%We could write this part as the paragraph in the introduction. Sth like:For most V\&L tasks, it has been shown that expensive V\&L pre-training has been instrumental for achieving high-performance. Meanwhile, for V\&L pre-training, [bascially say that people do not treat BERT initialization as a very important part]. We argue that in the medical domain, initializing the Transformer with a model pre-trained on in-domain data (even on a single modality), is crucial for achieving high performance.[Then add the current first paragraph]

%[Say thus we propose to initialize the Transformer with BlueBERT. You could say BlueBERT is an extention to BERT and keep the introduction of BERT short].
%\end{redcolor}

\section{Experiments}
\label{exp}
In this section, we evaluate BERTHop on the OpenI dataset and compare it with other existing models. To understand the effectiveness of the model designs, we also conduct detailed studies to verify the value of the visual encoder and the transformer initialization. Finally, we demonstrate a case study to show that BERTHop can effectively identify abnormal regions in CXR images. 
%In this section, we first introduce the dataset that is used for evaluation. Then, we discuss the hyper-parameters and other setting parameters in the training subsection and evaluation metric. Later on, we provide main results, discussions and analysis.
\subsection{Experiment setup}
\label{setup}
\paragraph{Dataset}

For CADx in CXR disease diagnosis, commonly used datasets include ChestX-ray14, MIMIC-CXR, and OpenI.
 In this paper, we focus on the OpenI dataset for which professional annotators labeled the data. OpenI comprises 3,996 reports and 8,121 associated images from 3,996 unique patients collected by Indiana University from multiple institutes. Its labels include 14 commonly occurring thoracic chest diseases, i.e., Atelectasis, Cardiomegaly, Effusion, Infiltration, Mass, Nodule, Pneumonia, Pneumothorax, Consolidation, Edema, Emphysema, Fibrosis, Pleural Thickening (PT), and Hernia. OpenI is a reliable choice for both training and evaluating V\&L models as it is annotated by experts (labels are not learned from text reports or images). The disadvantage of using OpenI for training is that it contains a small amount of training data which is a challenge for DNN models. We apply the same pre-processing as TieNet and obtain 3,684 image-text pairs.

We do not consider ChestX-ray14 and MIMIC-CXR for benchmarking because their labels are generated automatically from the images and/or associated reports. Specifically, ChestX-ray14 labels are mined using text process technique from the radiology reports, and MIMIC-CXR labels are generated using ChexPert\cite{irvin2019chexpert} and NegBio\cite{peng2018negbio} auto labelers. 
As their labels are machine-generated, evaluating the V\&L model on these datasets is not reliable. Therefore, we considered evaluation on OpenI to accurately compare the performance of BERTHop with human expert performance. 
%Therefore, they  are  not  valid  for  evaluating vision-and-language models. 
\begin{figure}
  \centering
  %\fbox{\rule[-.5cm]{0cm}{4cm} \rule[-.5cm]{4cm}{0cm}}
  %\includegraphics[width=10cm, height=6cm]{./fig/fig2_3}scale=1.5
  \includegraphics[width=\columnwidth]{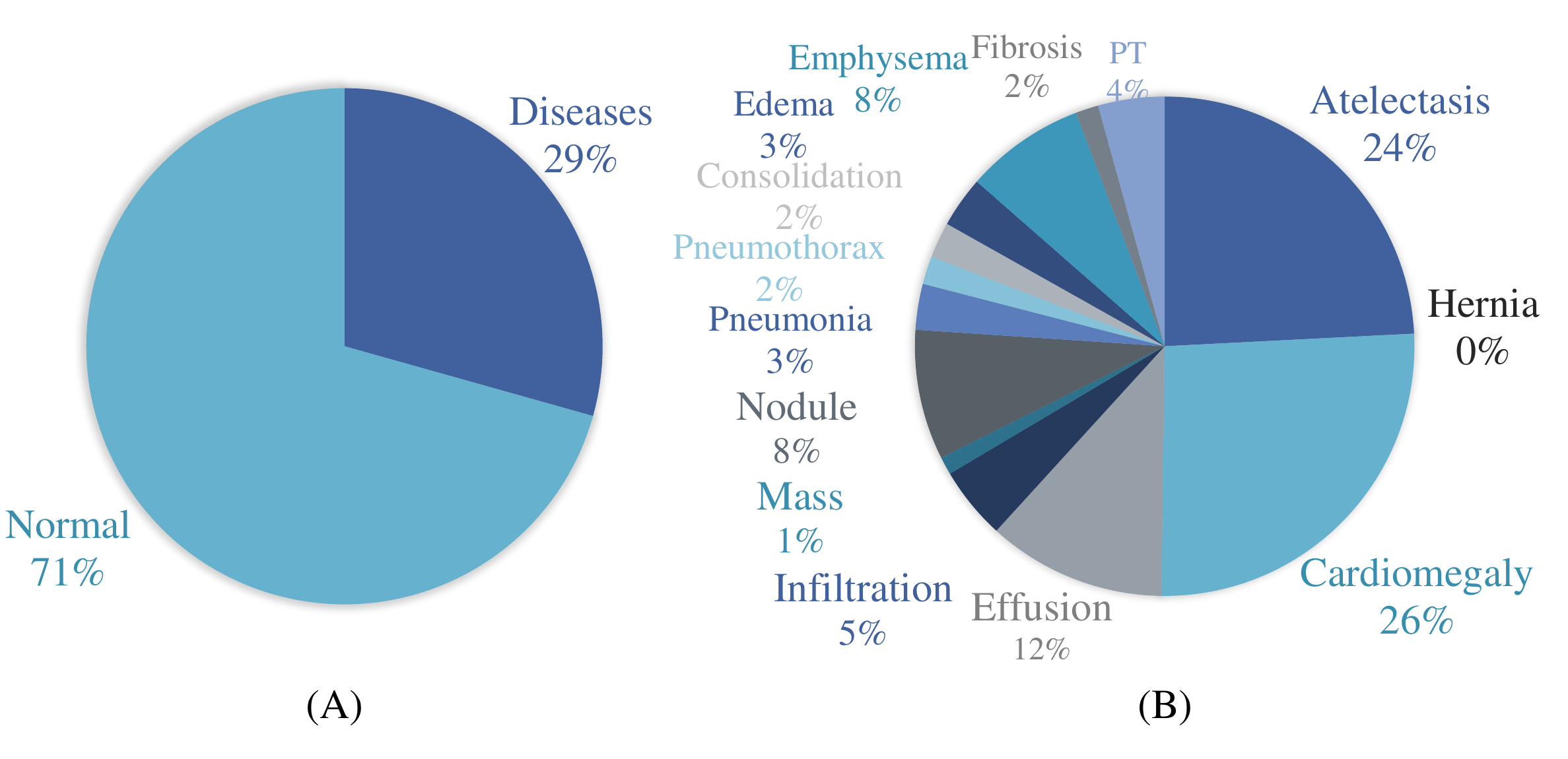}
  \caption{OpenI label statistics: (A) Percentage of normal and abnormal cases (B) Percentage of different diseases.}
\label{fig:stat}
\end{figure}

%Three chest x-ray datasets are commonly used for training and evaluating predictive models: Chest X-ray14, Mimic-CXR, and OpenI. Chest X-ray14 and mimic-CXR are suitable choices for training both vision and VL models. But as their labels are generated from the reports, they are not valid for evaluating V\&L models. On the other hand, OpenI is reliable for both training and evaluating V\&L models as it is annotated by experts (labels are not learned from text reports). The challenge with using OpenI for training is that it contains a small amount of training data which is a problem for DNNs. 
%These datasets are highly imbalanced and mostly contain normal cases. So, accuracy is not a suitable evaluation metric and proposed methods usually evaluate their models based on the ROC curve and AUC score.
%All three datasets contain 14 commonly occurring thoracic diseases. In this study, we focus on the OpenI dataset. The OpenI dataset comprises 3,996 reports and 8,121 associated images from 3,996 patients. We follow the same pre-processing steps that TieNet uses which provides 3,864 image-text pairs. 
%The Chest X-ray14 dataset contains 108,948 frontal images from 32,717 unique patients. The Mimic-CXR dataset is a large and the most recent publicly available chest x-ray dataset which includes 473,057 images and 206,563 reports collected from 63,478 patients. OpenI dataset comprises 3,996 reports and 8,121 associated images from 3,996 patients.
\paragraph{Model and training parameters} We first resize all images of OpenI to $206\times 206$ and apply the unsupervised feature learner, PixelHop++. We use a three-level PixelHop++ with the following hyper-parameters: $w = 3$, $d = 1$, $s = 1$, and $E = 0.00005$. Then, we apply PCA to its output channels and concatenate the generated vectors to form a set of $Q$ visual features of dimension D, i.e., $V = [v_1, v_2, ..., v_Q], v_i\in \mathbb{R}^D$. In BERTHop, $D$ is set to be 2048. In our experiments setup, $Q$ is equal to 15 but may vary depending on the size of the output channels of the PixelHop++ model and also the number of PCA components.
%form $Q = 15$ visual features of dimension 2048. 
%Then, we build text-image pairs which the text is encoded from BlueBERT tokenization. 

As for the transformer backbone, we use BlueBERT-Base (Uncased, PubMed+MIMIC-III) from Huggingface \cite{wolf2019huggingface}, a transformer library. %BlueBERT-Base is pre-trained on PubMed abstracts with more than 4 billion words in the biomedical domain and MIMIC-III with more than 500 million words in the clinical domain. 
Having the visual features from the visual encoder and text embedding, we train the transformer on the training set of OpenI with 2,912 image-text pairs. We use batch size = 18, learning rate = $1e-5$, max-seq-length = 128, and Stochastic Gradient Descent (SGD) as the optimizer with momentum = 0.9 and train it for 240 epochs. %The Receiver Operating Characteristic (ROC) curve andC (AUC) scores are reported in Fig. \ref{fig:ROC1} and table \ref{main_result} respectively.

%and expected the transformer to find the alignments.

\textbf{Evaluation metric} %Chest X-ray datasets are highly imbalanced mostly contain cases with no findings. We visualize the percentages of all disease labels for OpenI in Fig \ref{fig:stat}. The left graph shows the percentage of normal data vs all abnormal data (all diseases) and the right chart shows all diseases in one graph. Therefore, For disease diagnosis task, the accuracy metric is not suitable and the performance of the models need to be evaluated by ROC curve and Area Under the curve (AUC).
All mentioned datasets are highly imbalanced and mostly contain normal cases. Figure \ref{fig:stat} shows the percentages of different diseases compared with normal cases in OpenI. Therefore, evaluating models using metrics such as accuracy does not reflect model performance. Instead, we follow prior studies to evaluate models based on Receiver Operating Characteristic (ROC) and Area Under the ROC Curve (AUC) score.

\begin{table*}
\begin{center}
  \centering
  \begin{tabular}{ccccc}
  %\hline
    \toprule
    %\multicolumn{3}{c}{Part}                   \\
    %\cmidrule(r){2-5}
    & TNNT \cite{zhang2019text}  & TieNet$^*$ \cite{wang2018tienet} & VB w/ BUTD \cite{li2020comparison}  & BERTHop \\
    %\hline\hline
    \midrule
    Atelectasis & -  & 0.976  & 0.9247 & \textbf{0.9838}  \\
    Cardiomegaly & - & 0.962  & 0.9665 & \textbf{0.9896}     \\
    Effusion    & -  & \textbf{0.977}  & 0.9049 & 0.9432  \\
    Infiltration & - & 0.984  & 0.8867 & \textbf{0.9926}     \\
    Mass & -  & 0.903  & 0.6428 &       \textbf{0.9900} \\
    Nodule & -  & 0.960  & 0.8480 &     \textbf{0.9810} \\
    Pneumonia & -  & 0.994  & 0.8537 &  \textbf{0.9967} \\
    Pneumothorax & -  & 0.960  & 0.8931 & \textbf{1.0000} \\
    Consolidation & -  & \textbf{0.989}  & 0.7870 & 0.9671 \\
    Edema & -  & 0.995  & 0.9500 & \textbf{0.9987} \\
    Emphysema & -  & 0.868  & 0.8565 & \textbf{0.9971} \\
    Fibrosis & -  & 0.960  & 0.6274 & \textbf{0.9966} \\
    PT & -  & \textbf{0.953}  & 0.7612 & 0.9330 \\
    Hernia & -  & -  & - & - \\
    \midrule
    %\hline
    AVG & 0.854  & 0.965  & 0.8386 & \textbf{0.9823} \\
    %\hline
    
    \bottomrule
  \end{tabular}
  \vspace{5pt}
  \caption{The AUC thoracic diseases diagnosis comparison of our model with other three methods on OpenI. BERTHop significantly outperforms models trained with a similar amount of data (e.g. VB w/ BUTD). *TieNet is trained on a much larger dataset than BERTHop. }\label{main_result}
  \end{center}
\end{table*}

\subsection{Main results}
\label{subsec:main result}
\begin{figure}
  \centering
  %\fbox{\rule[-.5cm]{0cm}{4cm} \rule[-.5cm]{4cm}{0cm}}
  %\includegraphics[width=10cm, height=6cm]{./fig/fig2_3}scale=1.5
  \includegraphics[width=\columnwidth]{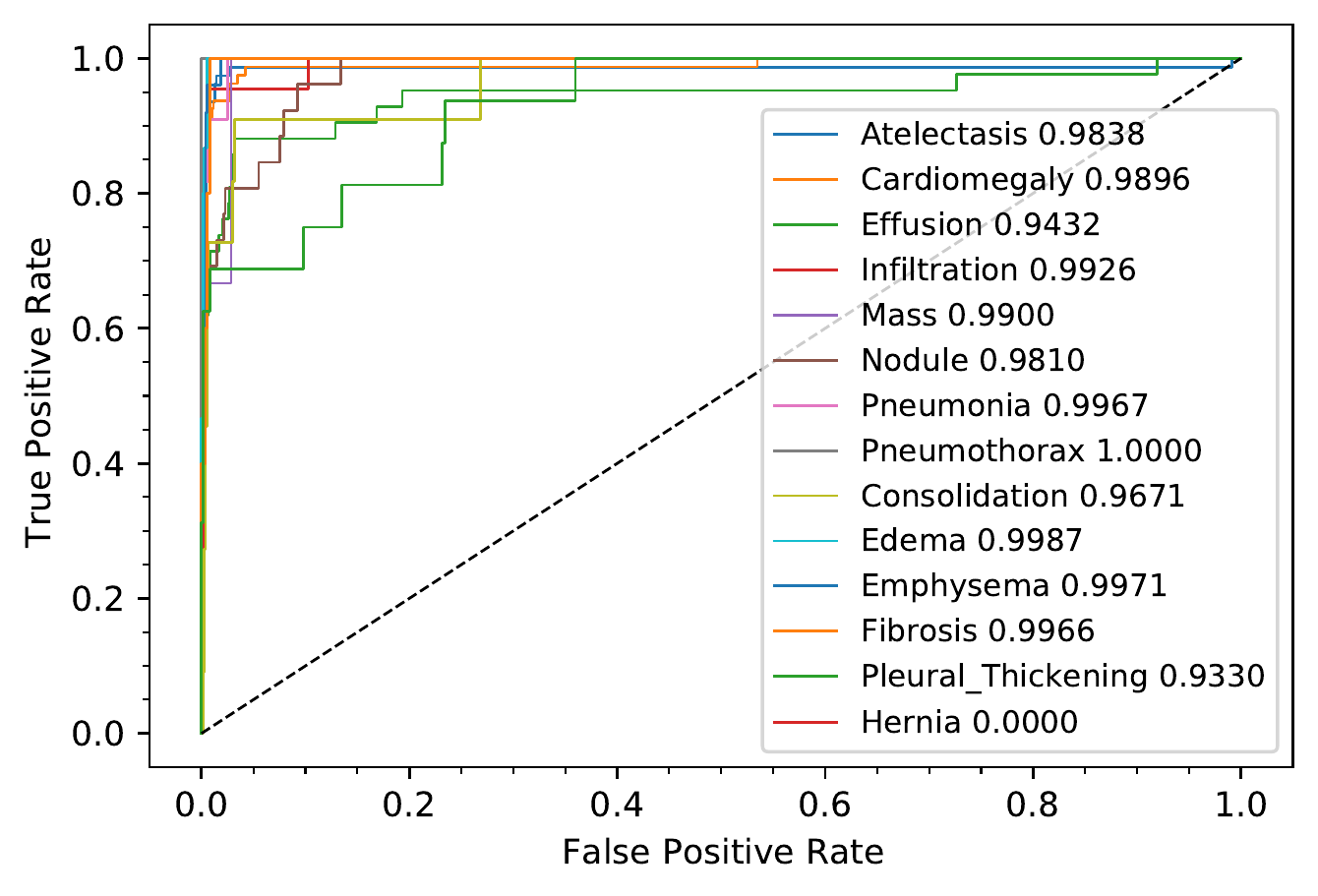}
  \caption{ROC curve of BERTHop for all 14 thoracic diseases.}
\label{fig:ROC1}
\end{figure}

%Having the visual features extracted from images using PixelHop++, we fine-tuned our model on the OpenI training dataset containing 2,912 image-text pairs and evaluated on its test set comprising 772 image-text pairs. 
We train BERTHop on the OpenI training dataset containing 2,912 image-text pairs and evaluate it on the corresponding test set comprising 772 image-text pairs. The ROC curve for each disease is plotted in Figure \ref{fig:ROC1}.

We compare BERTHop with the following approaches: 
\begin{itemize}
\item TNNT~\cite{zhang2019text}: a Text-giuded Nueral Network Training method. See the details in Section \ref{related}.
\item TieNET~\cite{wang2018tienet}: a CNN-RNN-based model. See the details in Section \ref{related}.
\item VB w/ BUTD~\cite{li2019visualbert,li2020comparison}: Fine-tuning the original VisualBERT.
\end{itemize}

%Figure \ref{fig:ROC1} and their AUC scores are computed. 
we evaluate all the models using the same AUC implementation in scikit-learn \cite{sklearn_api}. Table \ref{main_result} summarizes the performance of BERTHop compared with existing methods.

The results demonstrate that BERTHop outperforms SOTA (TieNet) in 11 out of 14 thoracic disease diagnoses and achieves an average AUC of 98.23\% which is 14.37\%, 12.83\%, and 1.73\% higher than VB w/ BUTD, TNNT, and TieNet, respectively. Note that TieNet has been trained on a much larger annotated dataset, i.e., the ChestX-ray14 dataset containing 108,948 training data while BERTHop is trained on only 2,912 case examples.

%?????????????????????
Regarding the VB w/ BUTD results, we re-evaluate the results based on the released code\footnote{\url{https://github.com/YIKUAN8/Transformers-VQA/blob/master/openI_VQA.ipynb}} from the original authors. However, we cannot reproduce the results reported in the paper even after contacting the authors. 
%We contacted the authors but we still cannot reproduce their results reported in the paper. We report the results in Table \ref{sample-table} based on their released code\footnote{\url{https://github.com/YIKUAN8/Transformers-VQA/blob/master/openI_VQA.ipynb}}.

\subsection{In-domain text pre-training}

We further investigate the influence of different transformer backbone initializations on model performance by pairing it with different visual encoders. The results are listed in Table \ref{BertBlue}.

First, we find that the proposed initialization with a model pre-trained on in-domain text corpora (BlueBERT) brings significant performance boosts when paired with PixelHop++. Initializing with BlueBERT gives a 6.46\% performance increase compared to initializing with BERT.

Second, when using BUTD, the model is less sensitive to the transformer initialization and the performance is generally low (from 83.09\% to 85.64\%). In contrast to other V\&L tasks \cite{li2019visualbert}, general-domain V\&L pre-training is not instrumental. 

The above findings suggest that for medical V\&L applications, in-domain single modality pre-training can bring larger performance improvement than using pre-trained V\&L models from the general domain, even though the latter is trained on a larger corpus. The relation and trade-off between single-modality pre-training and cross-modality pre-training are overlooked by previous works \cite{li2019visualbert} and we advocate for future research on this. 

%alleviates the burden of compute-intensive V\&L pre-training and shows the importance of in-domain single modality pre-training compared to general-domain cross-modal pre-training, which has not been discussed before.

%BERT has been pre-trained using two language tasks: 1) masked language modeling 2) next sentence prediction. Due to the shortage of clinical and biomedical words in BERT that may cause significant degradation of the generated embedding, we leverage BlueBERT and compare the results with BERT.  We compare BlueBERT with BERT by having visual features once from BUTD and once from BERTHop in Table \ref{BertBlue}. Based on these results, we can draw the conclusion that BlueBERT is more capable to be used in V\&L models in the medical domain. \harold{Move the introduction of BlueBERT to either method or experiment setup.}

\begin{table*}
\begin{center}
  \centering
  \begin{tabular}{c|ccc|cc}
  
    \toprule
    %\multicolumn{3}{c}{Part}                   \\
    %\cmidrule
    Visual Encoder& \multicolumn{3}{c|}{BUTD} & \multicolumn{2}{c}{PixelHop++} \\
    Transformer Backbone& VB & BERT  & BlueBERT & BERT & BlueBERT \\

    \midrule
    Atelectasis & 0.9247 & 0.8677  & 0.8866  & \textbf{0.9890} & 0.9838  \\
    Cardiomegaly & 0.9665 & 0.8877 & 0.8875  & 0.9772 & \textbf{0.9896}     \\
    Effusion    & 0.9049 & 0.8940  & 0.9120  & 0.9013 & \textbf{0.9432}  \\
    Mass & 0.6428 & 0.7365  & 0.7373  & 0.8886 & \textbf{0.9900} \\
    Consolidation & 0.7870 &  0.8766 & 0.8906  & 0.8949 & \textbf{0.9671} \\
    Emphysema & 0.8565 & 0.7313  & 0.8261  & 0.9641 & \textbf{0.9971} \\
    
    \midrule
    AVG & 0.8386 & 0.8309  & 0.8564  & 0.9177 & \textbf{0.9823} \\

    \bottomrule
  \end{tabular}
  \vspace{5pt}
  \caption{Effect of the transformer backbones when paired with different visual encoders. We find that when using BUTD features, the model becomes insensitive to the transformer initialization and the expensive V\&L pre-training brings little benefit compared to BERT initialization. When using PixelHop++, the model benefits significantly from initialization with BlueBERT, which is pre-trained on in-domain text corpora. }\label{BertBlue}
  \end{center}
\end{table*}

\subsection{Visual encoder} 
\label{ve}
To better understanding what visual encoder is suitable for medical applications, 
we compare three visual feature extraction methods (BUTD, ChexNet \cite{rajpurkar2017chexnet}, and PixelHop++). 
In particular, we replace the visual encoder of BERTHop with different visual encoders and report their performance. 
%by having the best transformer initialization from the previous analysis, i.e., BlueBERT.
BUTD extracts visual features from a Faster R-CNN pre-trained on Visual Genome, which is prevailing in recent V\&L models.
%While, to the best of our knowledge, no large-scale chest x-ray disease diagnosis dataset with bounding box annotation is available to be used for fine-tuning Faster R-CNN.
%The annotation should be very specific to the disease type. For example, cardiomegaly disease could be diagnose by looking at the width ratio of heart and chest that is difficult to be captured by BUTD that has been pre-trained in the general domain with lack of annotated data in the medical domain. 
ChexNet is a CNN-based method that is proposed for pneumonia disease detection. It is a 121-layer DenseNet \cite{huang2017densely} trained on the ChestX-ray14 dataset for pneumonia detection having all pneumonia cases labeled as positive examples and all other cases as negative examples. By modifying the loss function, it is also trained to classify
%as a multi-label classification model for 
all 14 thoracic diseases and achieved state-of-the-art among existing vision-only models, e.g., \cite{wang2017chestx}. To augment the data, it extracts 10 crops from the image (4 corners and one center and horizontally flipped version of them) and feeds it into the network to generate a feature vector of dimension 1024 for each of them. %So it generates 10 feature vector of size 1024 for each image. 
In order to make it compatible with our transformer framework, we apply a linear transformation that maps feature vectors of size 1024, generated by ChexNet, to 2048. We fine-tune ChexNet and train the parameters of the linear transformation on the OpenI dataset. 
%For a fair comparison, We initialize the transformer with BlueBERT in each experiment and show the result in Table \ref{tab:vision_table}.

The results in Table \ref{tab:vision_table} show that the visual encoder of BERTHop, PixelHop++, can extract richer features from the CXR images as it uses a data-efficient method capable of extracting image representations at different frequencies. Then, the transformer can highlight the most informative features from image-text data in an attention mechanism to make the final decision. In section \ref{abl}, we explore the visual encoder of BERTHop and its effectiveness to capture abnormality regions. 
%visualize PixelHop++ output channels for some disease types to illustrate its effectiveness better.

\begin{table}
\begin{center}
\centering
\begin{tabular}{cccc}
\toprule
%\multicolumn{3}{c}{Part}                   \\
%\cmidrule
& BUTD  & ChexNet & PixelHop++ \\
%\hline\hline
\midrule
Atelectasis & 0.8866  & 0.9787 & \textbf{0.9838}  \\
Cardiomegaly & 0.8875  & 0.9797 & \textbf{0.9896}     \\
Effusion     & 0.9120  & 0.8894 & \textbf{0.9432}  \\
Mass & 0.7373  & 0.7529 & \textbf{0.9900} \\
Consolidation & 0.8906  & 0.9000 & \textbf{0.9671} \\
Emphysema  & 0.8261  & 0.9067 & \textbf{0.9971} \\

\midrule

AVG & 0.8564  & 0.8798 & \textbf{0.9823} \\
%\hline

\bottomrule
\end{tabular}

\end{center}
\caption{Comparison betwee different visual encoders (BUTD, ChexNet, and PixelHop++) under the same transformer backbone of BlueBERT. PixelHop++ outperforms BUTD and even ChexNet, which is pre-trained on a large in-domain disease diagnosis dataset.}\label{tab:vision_table}
\end{table}

\section{Analysis}\label{abl}
\paragraph{Effectiveness of BERTHop with different dataset scales} To demonstrate the effectiveness of BERTHop on datasets of different scales and justify our designs, we conduct an experiment to compare BERTHop with its two variants: (1) In PH\_BERT, we replace BlueBERT with BERT. We compare BERTHop with PH\_BERT to show how a domain-specific BERT model helps to improve the performance in medical applications. (2) In BUTD\_BlueBERT, we replace the visual encoder PixelHop++ with the general visual encoder of BUTD. 

\begin{figure}
  \centering
  %\fbox{\rule[-.5cm]{0cm}{4cm} \rule[-.5cm]{4cm}{0cm}}
  %\includegraphics[width=10cm, height=6cm]{./fig/fig2_3}scale=1.5
  \includegraphics[width=\columnwidth]{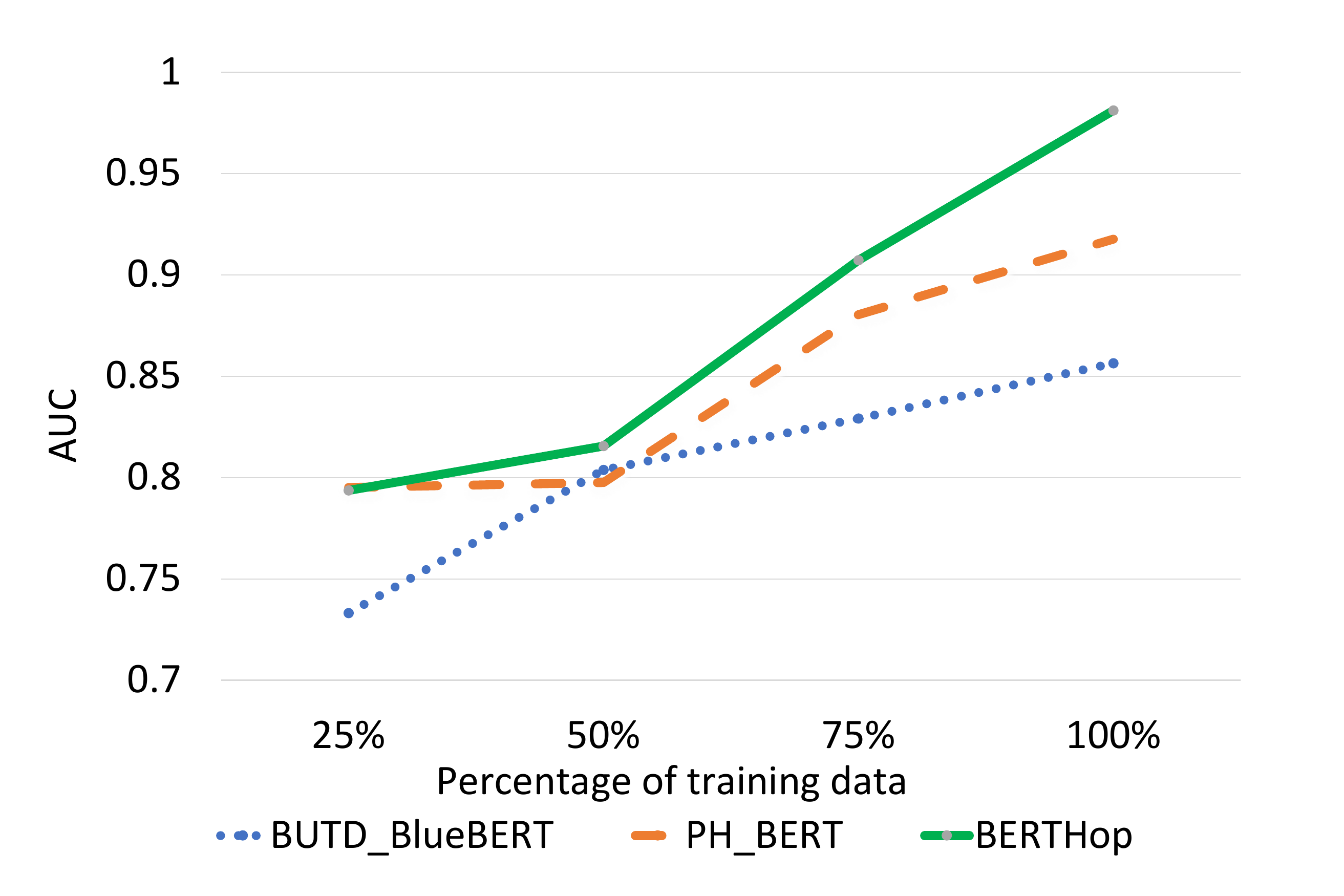}
  \caption{Avg AUC of three settings with different percentages of training data. BERTHop remains effective with different dataset scales.}
\label{fig:auc}
\end{figure}

We randomly select fractions of the training set of OpenI to train these three models and compare their performance on the entire test set of OpenI. Figure \ref{fig:auc} illustrates that the performance of BERTHop is consistently better than the other two settings.
%even when less annotated data is available.  %PixelHop++ is a data-efficient method for feature extraction and 
%the effectiveness of BERTHop  BERThop shows that its performance  in low source domains.

%To compare PixelHop++ with other visual feature extraction methods discussed above, we visualize where BUTD's visual encoder (Faster R-CNN) look at for feature extraction in Fig. Also, we asked a radiologist to annotated a few data examples for a qualitative analysis which are shown in right. BUTD seems 

\begin{figure}
  \centering
  %\fbox{\rule[-.5cm]{0cm}{4cm} \rule[-.5cm]{4cm}{0cm}}
  %\includegraphics[width=10cm, height=6cm]{./fig/fig2_3}scale=1.5
  \includegraphics[width=\columnwidth]{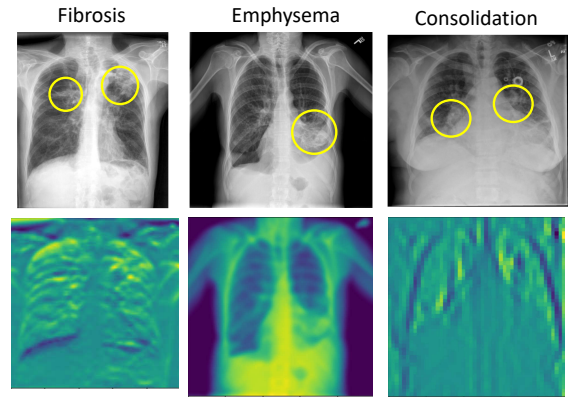}
  \caption{On the top, we mark the pathology regions annotated by two radiologists (the yellow circles and lines); on the bottom, we visualize the visual features from BERTHop (brighter colors means higher feature values). BERTHop can successfully highlight the abnormal regions identified by expert radiologists.}
\label{fig:abl}
\end{figure}

\paragraph{Visualize abnormal regions identified by BERTHop} We visualize PixelHop++ output channels of BERTHop to probe whether it can effectively capture abnormal regions in CXR images. In this study, we asked two radiologists to annotate pathology regions of a few examples related to different diseases.
As shown in Figure \ref{fig:abl}, some output channels can successfully highlight the abnormalities in CXR images. This is due to the fact that PixelHop++ extracts image representations at different frequencies which is beneficial for abnormality detection.
%To show where our visual encoder look at, we visualize its output in Fig. \ref{fig:abl}. the figure shows the visualization of our visual encoder along with a few annotated X-ray images. The visulization shows that the abnormal regions are highlighted in some channels of visual encoder which can be found by the transformer. For annotated data, we asked two radiologist to provide us a few annotated abnormalities.

%In Fig. , we show the ROC curve of BERTHop comapred with Fine-tuned VisualBERT. the curve shows the performance of our approach.

\section{Conclusion and future work}\label{con}
In this paper, we proposed a high-performance data-efficient V\&L model, BERTHop, for CXR disease diagnosis. We showed that BERTHop outperforms state-of-the-art while it is trained on a much smaller training set. Our studies verify the effectiveness of the visual feature extractor PixelHop++ and the transformer backbone initialization BlueBERT.
%We first discussed the previous V\&L models proposed for chest disease diagnosis and their challenges and how our proposed idea addresses the issues. 

For future research direction, we plan to study how anomaly detection techniques can be incorporated to further improve the performance of the model. As no large-scale annotated CXR dataset for anomaly detection is available, we may use weekly supervised techniques or knowledge transfer from similar tasks. We are also interested in how our proposed BERTHop model can help other biomedical tasks, e.g., COVID-19 disease diagnosis and radiology report generation. %Another future research direction is exploring the effectiveness of our visual encoder technique for other biomedical tasks, e.g., bladder cancer diagnosis.

{\small
\bibliographystyle{ieee_fullname}
\bibliography{egpaper_final}

\begin{thebibliography}{10}\itemsep=-1pt

\bibitem{abiyev2018deep}
Rahib~H Abiyev and Mohammad Khaleel~Sallam Ma’aitah.
\newblock Deep convolutional neural networks for chest diseases detection.
\newblock {\em Journal of healthcare engineering}, 2018, 2018.

\bibitem{allaouzi2019novel}
Imane Allaouzi and Mohamed~Ben Ahmed.
\newblock A novel approach for multi-label chest x-ray classification of common
  thorax diseases.
\newblock {\em IEEE Access}, 7:64279--64288, 2019.

\bibitem{anderson2018bottom}
Peter Anderson, Xiaodong He, Chris Buehler, Damien Teney, Mark Johnson, Stephen
  Gould, and Lei Zhang.
\newblock Bottom-up and top-down attention for image captioning and visual
  question answering.
\newblock In {\em Proceedings of the IEEE conference on computer vision and
  pattern recognition}, pages 6077--6086, 2018.

\bibitem{ayan2019diagnosis}
Enes Ayan and Halil~Murat {\"U}nver.
\newblock Diagnosis of pneumonia from chest x-ray images using deep learning.
\newblock In {\em 2019 Scientific Meeting on Electrical-Electronics \&
  Biomedical Engineering and Computer Science (EBBT)}, pages 1--5. Ieee, 2019.

\bibitem{sklearn_api}
Lars Buitinck, Gilles Louppe, Mathieu Blondel, Fabian Pedregosa, Andreas
  Mueller, Olivier Grisel, Vlad Niculae, Peter Prettenhofer, Alexandre
  Gramfort, Jaques Grobler, Robert Layton, Jake VanderPlas, Arnaud Joly, Brian
  Holt, and Ga{\"{e}}l Varoquaux.
\newblock {API} design for machine learning software: experiences from the
  scikit-learn project.
\newblock In {\em ECML PKDD Workshop: Languages for Data Mining and Machine
  Learning}, pages 108--122, 2013.

\bibitem{bustos2020padchest}
Aurelia Bustos, Antonio Pertusa, Jose-Maria Salinas, and Maria de~la
  Iglesia-Vay{\'a}.
\newblock Padchest: A large chest x-ray image dataset with multi-label
  annotated reports.
\newblock {\em Medical image analysis}, 66:101797, 2020.

\bibitem{chen2021defakehop}
Hong-Shuo Chen, Mozhdeh Rouhsedaghat, Hamza Ghani, Shuowen Hu, Suya You, and
  C.~C.~Jay Kuo.
\newblock Defakehop: A light-weight high-performance deepfake detector, 2021.

\bibitem{chen2020pixelhop++}
Yueru Chen, Mozhdeh Rouhsedaghat, Suya You, Raghuveer Rao, and C-C~Jay Kuo.
\newblock Pixelhop++: A small successive-subspace-learning-based (ssl-based)
  model for image classification.
\newblock In {\em 2020 IEEE International Conference on Image Processing
  (ICIP)}, pages 3294--3298. IEEE, 2020.

\bibitem{chen2020uniter}
Yen-Chun Chen, Linjie Li, Licheng Yu, Ahmed El~Kholy, Faisal Ahmed, Zhe Gan, Yu
  Cheng, and Jingjing Liu.
\newblock Uniter: Universal image-text representation learning.
\newblock In {\em European Conference on Computer Vision}, pages 104--120.
  Springer, 2020.

\bibitem{chou2020visual}
Shih-Han Chou, Wei-Lun Chao, Wei-Sheng Lai, Min Sun, and Ming-Hsuan Yang.
\newblock Visual question answering on 360deg images.
\newblock In {\em Proceedings of the IEEE/CVF Winter Conference on Applications
  of Computer Vision}, pages 1607--1616, 2020.

\bibitem{demner2016preparing}
Dina Demner-Fushman, Marc~D Kohli, Marc~B Rosenman, Sonya~E Shooshan, Laritza
  Rodriguez, Sameer Antani, George~R Thoma, and Clement~J McDonald.
\newblock Preparing a collection of radiology examinations for distribution and
  retrieval.
\newblock {\em Journal of the American Medical Informatics Association},
  23(2):304--310, 2016.

\bibitem{devlin2018bert}
Jacob Devlin, Ming-Wei Chang, Kenton Lee, and Kristina Toutanova.
\newblock Bert: Pre-training of deep bidirectional transformers for language
  understanding.
\newblock {\em arXiv preprint arXiv:1810.04805}, 2018.

\bibitem{fraser2019extracting}
Kathleen~C Fraser, Isar Nejadgholi, Berry De~Bruijn, Muqun Li, Astha LaPlante,
  and Khaldoun Zine~El Abidine.
\newblock Extracting umls concepts from medical text using general and
  domain-specific deep learning models.
\newblock {\em arXiv preprint arXiv:1910.01274}, 2019.

\bibitem{giger2008computer}
Maryellen~L Giger and Kenji Suzuki.
\newblock Computer-aided diagnosis.
\newblock In {\em Biomedical information technology}, pages 359--XXII.
  Elsevier, 2008.

\bibitem{gu2020domain}
Yu Gu, Robert Tinn, Hao Cheng, Michael Lucas, Naoto Usuyama, Xiaodong Liu,
  Tristan Naumann, Jianfeng Gao, and Hoifung Poon.
\newblock Domain-specific language model pretraining for biomedical natural
  language processing.
\newblock {\em arXiv preprint arXiv:2007.15779}, 2020.

\bibitem{huang2017densely}
Gao Huang, Zhuang Liu, Laurens Van Der~Maaten, and Kilian~Q Weinberger.
\newblock Densely connected convolutional networks.
\newblock In {\em Proceedings of the IEEE conference on computer vision and
  pattern recognition}, pages 4700--4708, 2017.

\bibitem{irvin2019chexpert}
Jeremy Irvin, Pranav Rajpurkar, Michael Ko, Yifan Yu, Silviana Ciurea-Ilcus,
  Chris Chute, Henrik Marklund, Behzad Haghgoo, Robyn Ball, Katie Shpanskaya,
  et~al.
\newblock Chexpert: A large chest radiograph dataset with uncertainty labels
  and expert comparison.
\newblock In {\em Proceedings of the AAAI Conference on Artificial
  Intelligence}, volume~33, pages 590--597, 2019.

\bibitem{jain2021visualchexbert}
Saahil Jain, Akshay Smit, Steven~QH Truong, Chanh~DT Nguyen, Minh-Thanh Huynh,
  Mudit Jain, Victoria~A Young, Andrew~Y Ng, Matthew~P Lungren, and Pranav
  Rajpurkar.
\newblock Visualchexbert: Addressing the discrepancy between radiology report
  labels and image labels.
\newblock {\em arXiv preprint arXiv:2102.11467}, 2021.

\bibitem{johnson2019mimic}
Alistair~EW Johnson, Tom~J Pollard, Nathaniel~R Greenbaum, Matthew~P Lungren,
  Chih-ying Deng, Yifan Peng, Zhiyong Lu, Roger~G Mark, Seth~J Berkowitz, and
  Steven Horng.
\newblock Mimic-cxr-jpg, a large publicly available database of labeled chest
  radiographs.
\newblock {\em arXiv preprint arXiv:1901.07042}, 2019.

\bibitem{krishna2017visual}
Ranjay Krishna, Yuke Zhu, Oliver Groth, Justin Johnson, Kenji Hata, Joshua
  Kravitz, Stephanie Chen, Yannis Kalantidis, Li-Jia Li, David~A Shamma, et~al.
\newblock Visual genome: Connecting language and vision using crowdsourced
  dense image annotations.
\newblock {\em International journal of computer vision}, 123(1):32--73, 2017.

\bibitem{kuo2019interpretable}
C-C~Jay Kuo, Min Zhang, Siyang Li, Jiali Duan, and Yueru Chen.
\newblock Interpretable convolutional neural networks via feedforward design.
\newblock {\em Journal of Visual Communication and Image Representation},
  60:346--359, 2019.

\bibitem{li2019visualbert}
Liunian~Harold Li, Mark Yatskar, Da Yin, Cho-Jui Hsieh, and Kai-Wei Chang.
\newblock Visualbert: A simple and performant baseline for vision and language.
\newblock {\em arXiv preprint arXiv:1908.03557}, 2019.

\bibitem{li2020comparison}
Yikuan Li, Hanyin Wang, and Yuan Luo.
\newblock A comparison of pre-trained vision-and-language models for multimodal
  representation learning across medical images and reports.
\newblock In {\em 2020 IEEE International Conference on Bioinformatics and
  Biomedicine (BIBM)}, pages 1999--2004. IEEE, 2020.

\bibitem{liu2021voxelhop}
Xiaofeng Liu, Fangxu Xing, Chao Yang, C-C~Jay Kuo, Suma Babu, Georges~El
  Fakhri, Thomas Jenkins, and Jonghye Woo.
\newblock Voxelhop: Successive subspace learning for als disease classification
  using structural mri.
\newblock {\em arXiv preprint arXiv:2101.05131}, 2021.

\bibitem{lu2019vilbert}
Jiasen Lu, Dhruv Batra, Devi Parikh, and Stefan Lee.
\newblock Vilbert: Pretraining task-agnostic visiolinguistic representations
  for vision-and-language tasks.
\newblock {\em arXiv preprint arXiv:1908.02265}, 2019.

\bibitem{lu202012}
Jiasen Lu, Vedanuj Goswami, Marcus Rohrbach, Devi Parikh, and Stefan Lee.
\newblock 12-in-1: Multi-task vision and language representation learning.
\newblock In {\em Proceedings of the IEEE/CVF Conference on Computer Vision and
  Pattern Recognition}, pages 10437--10446, 2020.

\bibitem{peng2018negbio}
Yifan Peng, Xiaosong Wang, Le Lu, Mohammadhadi Bagheri, Ronald Summers, and
  Zhiyong Lu.
\newblock Negbio: a high-performance tool for negation and uncertainty
  detection in radiology reports.
\newblock {\em AMIA Summits on Translational Science Proceedings}, 2018:188,
  2018.

\bibitem{peng2019transfer}
Yifan Peng, Shankai Yan, and Zhiyong Lu.
\newblock Transfer learning in biomedical natural language processing: an
  evaluation of bert and elmo on ten benchmarking datasets.
\newblock {\em arXiv preprint arXiv:1906.05474}, 2019.

\bibitem{peters2018deep}
Matthew~E Peters, Mark Neumann, Mohit Iyyer, Matt Gardner, Christopher Clark,
  Kenton Lee, and Luke Zettlemoyer.
\newblock Deep contextualized word representations.
\newblock {\em arXiv preprint arXiv:1802.05365}, 2018.

\bibitem{rajpurkar2017chexnet}
Pranav Rajpurkar, Jeremy Irvin, Kaylie Zhu, Brandon Yang, Hershel Mehta, Tony
  Duan, Daisy Ding, Aarti Bagul, Curtis Langlotz, Katie Shpanskaya, et~al.
\newblock Chexnet: Radiologist-level pneumonia detection on chest x-rays with
  deep learning.
\newblock {\em arXiv preprint arXiv:1711.05225}, 2017.

\bibitem{ren2016faster}
Shaoqing Ren, Kaiming He, Ross Girshick, and Jian Sun.
\newblock Faster r-cnn: towards real-time object detection with region proposal
  networks.
\newblock {\em IEEE transactions on pattern analysis and machine intelligence},
  39(6):1137--1149, 2016.

\bibitem{rouhsedaghat2021successive}
Mozhdeh Rouhsedaghat, Masoud Monajatipoor, Zohreh Azizi, and C-C~Jay Kuo.
\newblock Successive subspace learning: An overview.
\newblock {\em arXiv preprint arXiv:2103.00121}, 2021.

\bibitem{rouhsedaghat2020facehop}
Mozhdeh Rouhsedaghat, Yifan Wang, Xiou Ge, Shuowen Hu, Suya You, and C-C~Jay
  Kuo.
\newblock Facehop: A light-weight low-resolution face gender classification
  method.
\newblock {\em arXiv preprint arXiv:2007.09510}, 2020.

\bibitem{rouhsedaghat2020low}
Mozhdeh Rouhsedaghat, Yifan Wang, Shuowen Hu, Suya You, and C-C~Jay Kuo.
\newblock Low-resolution face recognition in resource-constrained environments.
\newblock {\em arXiv preprint arXiv:2011.11674}, 2020.

\bibitem{shin2016deep}
Hoo-Chang Shin, Holger~R Roth, Mingchen Gao, Le Lu, Ziyue Xu, Isabella Nogues,
  Jianhua Yao, Daniel Mollura, and Ronald~M Summers.
\newblock Deep convolutional neural networks for computer-aided detection: Cnn
  architectures, dataset characteristics and transfer learning.
\newblock {\em IEEE transactions on medical imaging}, 35(5):1285--1298, 2016.

\bibitem{su2019vl}
Weijie Su, Xizhou Zhu, Yue Cao, Bin Li, Lewei Lu, Furu Wei, and Jifeng Dai.
\newblock Vl-bert: Pre-training of generic visual-linguistic representations.
\newblock {\em arXiv preprint arXiv:1908.08530}, 2019.

\bibitem{tan2019lxmert}
Hao Tan and Mohit Bansal.
\newblock Lxmert: Learning cross-modality encoder representations from
  transformers.
\newblock {\em arXiv preprint arXiv:1908.07490}, 2019.

\bibitem{wada2020pre}
Shoya Wada, Toshihiro Takeda, Shiro Manabe, Shozo Konishi, Jun Kamohara, and
  Yasushi Matsumura.
\newblock A pre-training technique to localize medical bert and enhance
  biobert.
\newblock {\em arXiv preprint arXiv:2005.07202}, 2020.

\bibitem{wang2017chestx}
Xiaosong Wang, Yifan Peng, Le Lu, Zhiyong Lu, Mohammadhadi Bagheri, and
  Ronald~M Summers.
\newblock Chestx-ray8: Hospital-scale chest x-ray database and benchmarks on
  weakly-supervised classification and localization of common thorax diseases.
\newblock In {\em Proceedings of the IEEE conference on computer vision and
  pattern recognition}, pages 2097--2106, 2017.

\bibitem{wang2018tienet}
Xiaosong Wang, Yifan Peng, Le Lu, Zhiyong Lu, and Ronald~M Summers.
\newblock Tienet: Text-image embedding network for common thorax disease
  classification and reporting in chest x-rays.
\newblock In {\em Proceedings of the IEEE conference on computer vision and
  pattern recognition}, pages 9049--9058, 2018.

\bibitem{wolf2019huggingface}
Thomas Wolf, Lysandre Debut, Victor Sanh, Julien Chaumond, Clement Delangue,
  Anthony Moi, Pierric Cistac, Tim Rault, R{\'e}mi Louf, Morgan Funtowicz,
  et~al.
\newblock Huggingface's transformers: State-of-the-art natural language
  processing.
\newblock {\em arXiv preprint arXiv:1910.03771}, 2019.

\bibitem{zhang2019text}
Zizhao Zhang, Pingjun Chen, Xiaoshuang Shi, and Lin Yang.
\newblock Text-guided neural network training for image recognition in natural
  scenes and medicine.
\newblock {\em IEEE transactions on pattern analysis and machine intelligence},
  2019.

\bibitem{zhou2020unified}
Luowei Zhou, Hamid Palangi, Lei Zhang, Houdong Hu, Jason Corso, and Jianfeng
  Gao.
\newblock Unified vision-language pre-training for image captioning and vqa.
\newblock In {\em Proceedings of the AAAI Conference on Artificial
  Intelligence}, volume~34, pages 13041--13049, 2020.

\end{thebibliography}
}

\end{document}